\documentclass[10pt, conference, letterpaper]{IEEEtran}
\IEEEoverridecommandlockouts
\usepackage{cite}
\usepackage{amsmath,amssymb,amsfonts}
\usepackage{algorithmic}
\usepackage{graphicx}
\usepackage{textcomp}
\usepackage{xcolor}
\usepackage{blkarray}
\usepackage{slashbox}
\usepackage{placeins}
\usepackage{soul}
\usepackage{caption,subfloat,subfig}
\def\BibTeX{{\rm B\kern-.05em{\sc i\kern-.025em b}\kern-.08em
    T\kern-.1667em\lower.7ex\hbox{E}\kern-.125emX}}

\definecolor{darkorchid}{rgb}{0.6, 0.2, 0.8}

\definecolor{blue}{rgb}{0, 0, 1}

\begin{document}

\title{Towards Adaptive IMFs - Generalization of utility functions in Multi-Agent Frameworks \thanks{Accepted as short-paper in Netsoft-2024 conference, St. Louis, USA}}

\author{\IEEEauthorblockN{Kaushik Dey and Satheesh K. Perepu}
\IEEEauthorblockA{\textit{Ericsson Research (Artificial Intelligence), India}\\
\{deykaushik, perepu.satheesh.kumar\}@ericsson.com}
\and
\IEEEauthorblockN{Abir Das}
\IEEEauthorblockA{\textit{Indian Institute of Technology, Kharagpur, India} \\
abir@cse.iitkgp.ac.in}
\and
\IEEEauthorblockN{Pallab Dasgupta}
\IEEEauthorblockA{\textit{Synopsys Inc., USA}\\
pallabd@synopsys.com}
}

\maketitle

\begin{abstract}
Intent Management Function (IMF) is an integral part of future-generation networks. In recent years, there has been some work on AI-based IMFs that can handle conflicting intents and prioritize the global objective based on apriori definition of the utility function and accorded priorities for competing intents. Some of the earlier works use Multi-Agent Reinforcement Learning (MARL) techniques with AdHoc Teaming (AHT) approaches for efficient conflict handling in IMF. However, the success of such frameworks in real-life scenarios requires them to be flexible to business situations. The intent priorities can change and the utility function, which measures the extent of intent fulfilment, may also vary in definition. This paper proposes a novel mechanism whereby the IMF can generalize to different forms of utility functions and change of intent priorities at run-time without additional training. Such generalization ability, without additional training requirements, would help to deploy IMF in live networks where customer intents and priorities change frequently. Results on the network emulator demonstrate the efficacy of the approach, scalability for new intents, outperforming existing techniques that require additional training to achieve the same degree of flexibility thereby saving cost, and increasing efficiency and adaptability.
\end{abstract}


\section{Introduction}
\label{sec:intro}

Networks, in future, are expected to be driven by multiple intents originating from various customers. A single intent can contain multiple expectations on different KPI's. An example of an intent with one expectation is ``90\% of users accessing conversational video service should get Quality of Experience (QoE) $\geq$ 4" \cite{web:tmforum}. In some cases, intents conflict with each other as the available resources are limited. Intent driven networks in 6G may be orchestrated by AI-enabled IMFs where each IMF can be responsible for a domain or part of the domain \cite{report:Ericsson,paper:intent_review}. Here each IMF needs to orchestrate multiple systems to fulfil the intents or sub-intents allocated to it. Each such system can actually be an autonomous group of agents acting as a cohesive entity under a cooperative framework, like MARL~\cite{paper:globecom}. The IMF needs to manage more than one such MARL systems and in such a case, may employ a supervisor to orchestrate different MARL systems. The supervisor agent receives the intent for that domain and decomposes the intent into sub-goals for these MARL systems. The decomposition and allocation of the sub-goals to system of agents can be done by an AI enabled algorithm which learns and adapts by observing the performance of all the agents in its purview. The supervisor agent through sub-goals indirectly induces coordination among these MARL systems and thereby enables fulfillment of intents\cite{paper:dey2023goals} for the IMF.

Elaborating on such system of agents, each agent in a system may be responsible for optimizing a parameter(s) for a service. Examples of such services could be Conversational Video(CV), Ultra Reliable Low Latency Communication (URLLC), and Massive IoT (mIoT). Such IMFs have been defined in \cite{MEHMOOD2023109477}, \cite{report:Ericsson} and \cite{paper:globecom}. In 6G networks, the number of such services could expand further with development of new applications, which necessitates development of AI-based IMF frameworks which can autonomously fulfill multiple intents. 


As per TM forum standard each intent/expectation is associated with an utility function \cite{web:tmforum}. The utility value (function value) of an intent is defined as the relative importance of satisfying all expectation/s of an intent. When multiple intents, competing in resources, are present in the system, the utilities of all such intents may be expressed as a global Utility/ Objective function which may be expressed in following form
\begin{equation}
Z(\textbf{x}) = P_1 \cdot f_1(T_1-x_1) + P_2 \cdot f_2(T_2-x_2) + \cdots  P_n \cdot f_n(T_n-x_n)
\label{eq:eq_obj}
\end{equation}

where $f_{i}(.), \; i = 1,\cdots n, $ is the utility function for expectation $i$, $P_i$ the priority accorded, $T_i$ the expectation on associated KPI, $x_i$ the current value of KPI of the expectation on the intent. 



The agents in IMF perform actions such that the function $Z(x)$ is either maximized or minimized depending on the context. Current literature proposes using MARL\cite{paper:globecom} or model-based Machine Reasoning approaches\cite{etsi-zsm-closedloops} to optimize various intents in IMF through the \textit{pre-defined} utility function. However, if the resources are limited, all the expectations in the intents may still not be fulfilled. In such a case intents may often need to be prioritized in run-time to specify their relative importance. The supervisor agent then orchestrates the lower-level (MARL) agents to optimize the global utility function as per the user-defined priorities. As network resources become limited, such a mechanism to achieve appropriate trade-offs using prioritized intents in IMF may be the key to competitive differentiation in future 6G networks.

\begin{figure*}[ht]
    \centering
    \includegraphics[scale=0.5]{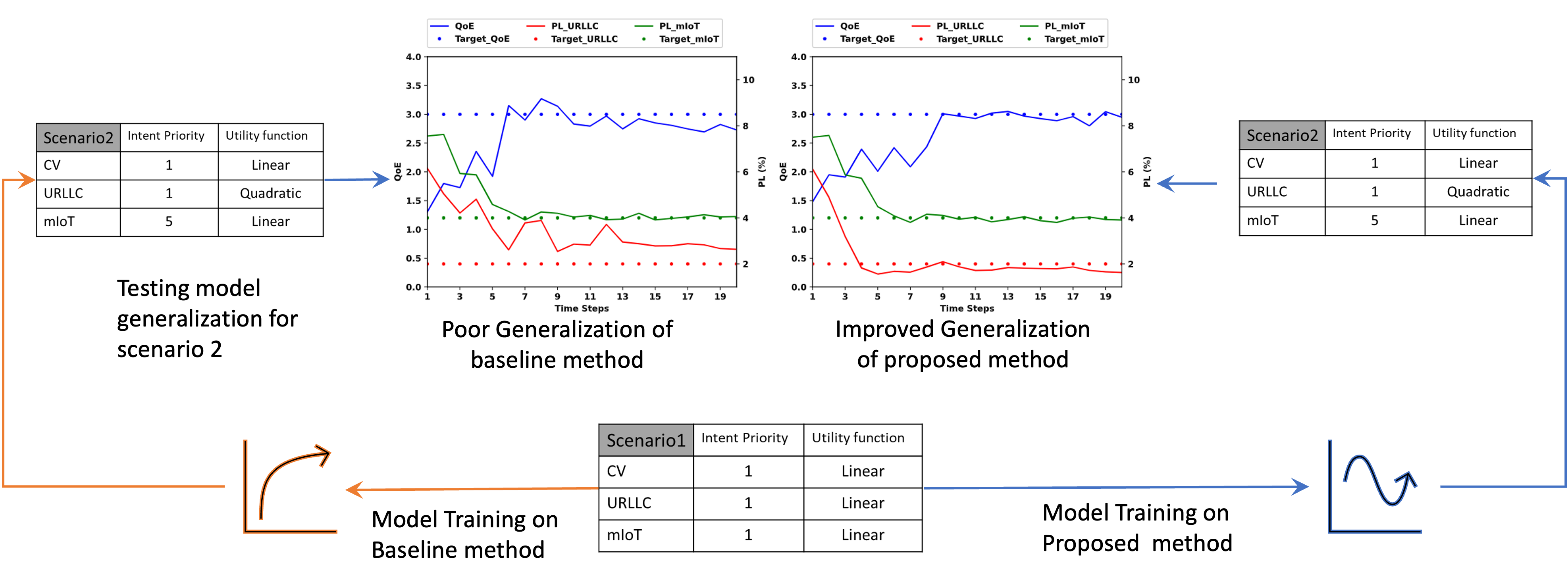}
    \caption{Block diagram depicting performance of baseline method (Fig 1a) and thereby need for improved convergence for unseen situations i.e when intent priority or function form changes. The plot in 1b depicts the significant improvement in convergence for the same unseen situations i.e generalization abilities when our proposed method is leveraged.}
    \label{fig:intropic}
\end{figure*}

    

For the case of MARL, which is of particular interest for our work, the values obtained from the objective function \eqref{eq:eq_obj}, at each time step, needs to be incorporated in the reward signal during the training phase. The reward signal is used to train a RL policy for the supervisor agent. The supervisor leverages its RL policy to propose auxiliary goals for the constituent agents of the IMF. This approach requires the form (definition) of $Z(x)$ and $P_i$ values to be specified during the training phase to arrive at an optimal RL policy. Implementing such an approach, using existing methods, has limitations and challenges for live networks as summarized below: 
\begin{enumerate}
    \item Priority($P_i$) is dynamic as it depends on the business criticality of the intent at that time instant. 
    The priority often varies across services and an IMF needs to adapt to changing priorities during the execution phase.
    \item The utility function definition can change, for example from linear to quadratic or logarithmic forms. For example if a stricter requirement on fulfillment of a specific intent is required, a quadratic function might be preferred over a linear form as explained subsequently. 
    \item Above changes may be initiated by operator during execution in live networks and without the opportunity for additional training of the constituent models.
\end{enumerate}

Existing literature provides limited solutions for IMFs to perform optimally in face of such challenges as they require a re-training phase for changes to utility functions or priorities. In case of autonomous IMF frameworks, in order to adjust to such changes at run-time, the trained models would need to rewire the collaboration and calibrate the trade-off behaviors to optimize the intents, specially in case where resources are limited. IMF frameworks in current literature are not designed to deliver this level of flexibility without additional training. 
However such features are essential for implementing IMF in real life networks for reasons explained below:

\begin{enumerate}
    \item In a shared slice running multiple services like URLLC, mIoT and CV, consider a resource constrained scenario when all intents in the slice cannot be fulfilled and hence prioritization is necessary. For example, an operator might onboard various services in a slice but an URLLC service serving a low latency manufacturing application in factory might need to be prioritized and guaranteed from performance aspect during a span of the day. A streaming service may need similar prioritization during a game.  Such prioritization changes through $P_i$ values, should ideally induce the RL agents to re-calibrate the resource attribution patterns learned during training phase. But given the agents are trained to collaborate based a given set of priorities, they have limitations to adjust when priorities change during execution phase.
    \item The utility function form which measures the intent fulfillment may also be leveraged as an even more powerful tool for prioritization. A quadratic utility function imposes more stringent adherence to expectations while a logarithmic is more tolerant to deviations. The former is also expected to lead to faster convergence during execution as compared to later as evident from equation\ref{eq:eq_obj}. Hence the operator may leverage it to drive differentiated behaviors for intents based on needs for monetization by changing the function form in run-time.
\end{enumerate}
The challenge is illustrated in fig. \ref{fig:intropic} where we train the existing baseline model from scratch, for \textit{scenario-1}, using linear utility functions(eg. $P_i|(T_i-x_i)|$) and $P_i=1$ for all intents.
The model obtained herein is tested for \textit{scenario-2}, with CV({$P_1=1$, Linear}), URLLC({$P_2=1$, Quadratic}) and mIoT({$P_3=5$, linear}).
Since the existing method, which is our baseline for comparison, does not have the intelligence or provision to generalize, it shows degraded performance when tested under unseen circumstances. This depicts the principal challenge which we propose to solve. Our method, explained in subsequent sections, is able to ingest and adapt the changed objective functions and priorities during run-time without any additional training. As a result we see a significant improvement in results whereby URLLC intent achieves a packet loss less than target value, the CV intent achieves higher QoE than target and mIoT intent also achieves the desired packet loss. The URLLC intent under quadratic utility also reaches it target faster than the mIot and CV which are under linear utility function further confirming the adaptation to changed utility functions.  

For purposes of our work we have choose  an AHT-based approach combined \cite{paper:m3rl} with a MARL-based framework \cite{rashid2020qmix}. Integrating the two approaches can provide a scalable and performant IMF which can handle multiple intents as seen in \cite{paper:dey2023goals}. We have used the same network simulator as in \cite{paper:globecom} and \cite{paper:dey2023goals} for purposes of benchmarking with existing work.

The IMF, implemented using a MARL-based framework \cite{paper:dey2023goals}, in a network domain may consist of more than one subordinate MARL systems as depicted in fig. \ref{fig:ICC_BD}. Each MARL system may optimize KPIs for multiple intents and modifies a specific control parameter. Such systems of RL agents might arrive pre-trained and is often unfamiliar to existing systems. Hence to coordinate systems of RL agents or groups of MARL agents, supervisor agent trained using AHT approaches, can be effective as seen in \cite{paper:dey2023goals}. Our work is a continuation from the above mentioned literature whereby we extend the IMF implementation, by \textit{introducing a generalization mechanism for the utility function} so that the priority of intents or form of the function can be changed per needs of the business users at run-time without any additional training. The novel outcome requires a simple modification to the existing framework in terms of a DUN network along with feature engineering, which adds minimal computational requirement to the training process. Finally we scale the framework for new expectations on the intents adding \textit{latency} and \textit{power consumption} with new action spaces and prove that our generalization approach scales for increased dimension and heterogeneity of actions. Our contributions are summarized as follows and are specially applicable to resource constrained scenarios:
\begin{enumerate}
    \item The proposed method adapts to dynamic changes to assigned priorities of intents at run-time.
    \item The proposed method also enables the form of the utility function to be altered in run time which changes the length of execution episode and can help to selectively achieve a faster goal attainment for chosen intents.
    \item The method is scalable and generalizable to increased number of expectations as it can accommodate increase in dimension  and heterogeneity in action space
    \item It is a lightweight and simple addition to existing state of art with little computational overhead while making the IMF adaptive and flexible to business requirements.

\end{enumerate}

As far as we know, the work represents the first of its kind towards generalizing IMFs handling complex multi-agent systems to \textit{change of utility functions and intent priorities} in run time to address dynamic priorities of the operator.



\section{Background}
\label{sec:BGD}

\subsection{Literature Review}
\label{subsec:lit}
While there exists some literature on AI-based orchestration of services\cite{paper:AI_Orchestration} and service assurance\cite{paper:globecom} in IMFs, the challenge to generalize utility function definitions and intent priorities in IMF has been unexplored till date \cite{paper:globecom,paper:5g_orchestration}. 
Hence this section also explores the literature on flexibility of Utility Functions in context of Reinforcement Learning frameworks and not necessarily in context of IMFs only. 

Multi-Objective Reinforcement Learning (MORL) has been a fairly well studied topic but principally in context of single agents \cite{paper:MORL1,paper:MORL2}. In MORL, agent tries to learn a policy which optimizes more than one conflicting objectives. The two categories of MORL approaches and their respective challenges are listed below:
\begin{enumerate}
    \item Converting the multiple conflicting objectives to a single objective function: This approach may result in an averaged policy, non-optimal in some circumstances.
    \item Devising multiple policies on a pre-defined Pareto frontier: The need for apriori definition of pareto frontier may lead to lack of scalability and flexibility.
\end{enumerate}
There has also been some work on Multi Objective MARL (MOMARL) \cite{paper:Moffaert,Mannion2018RewardSF, paper:MOMIX} where multiple agents try to optimize a combination of conflicting objectives. \textit{All these methods assume that the global objective function remains unchanged during execution} and hence cannot be applicable to our case as the global utility function changes during execution. 


\begin{figure*}[ht]
    \centering
    \includegraphics[scale=0.28]{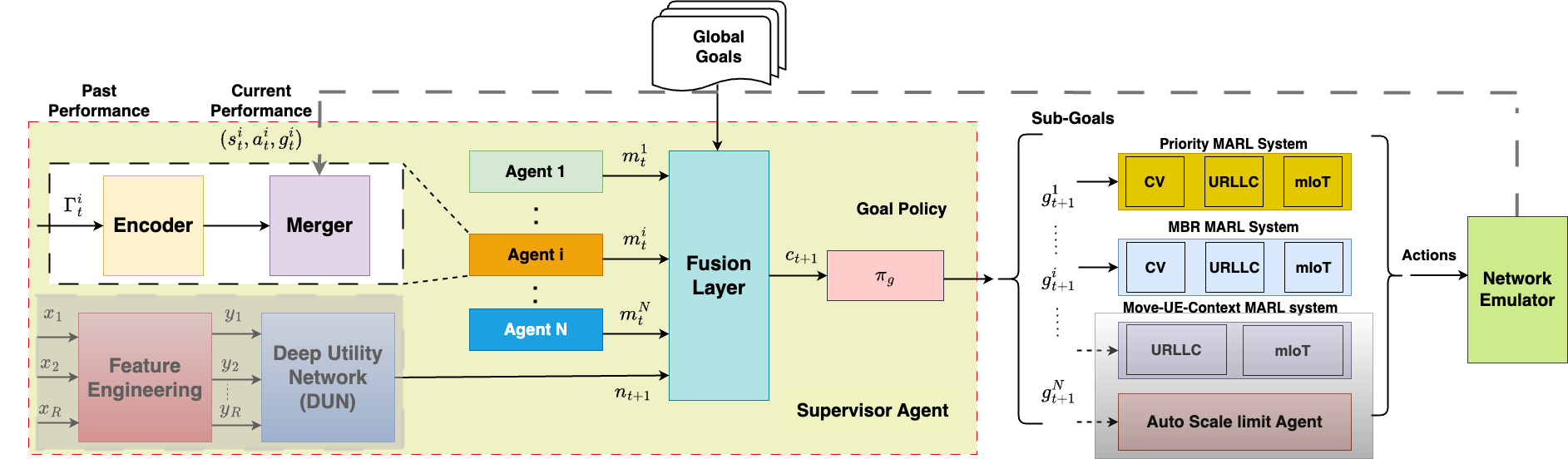}
    \caption{Block diagram of the proposed methodology. The supervisor agent assigns sub-goals to the lower-level agents by observing current and past performance, global goals (derived from intents) and utility context $n_{t+1}$. The observed KPI values ($x_i$ for agent $i$) for all agents are send to feature engineering block (as per \eqref{eqn:feature}) and extracted features($y_i$) to DUN to create utility context $n_{t+1}$. The four types of lower-level agents used in this work are: The priority MARL systems of 3 agents which controls unique packet priorities for CV, URLLC and mIoT. The MBR MARL system which changes Max Bit rate. The third corresponds to Move-UE-context system which contains two agents (one for URLLC and another for mIoT). The fourth one is a single RL agent. The areas with semi-transparent(gray) boxes are our contributions when compared to existing literature.}
    \label{fig:ICC_BD}
\end{figure*}

\subsection{Related Work}
\label{subsec:related}

For the purposes of this work we consider an IMF as defined in \cite{paper:globecom} with three distinct service types - CV, URLLC and mIoT. Each service exhibits an unique network traffic characteristic. Our goal is to manage the intents across all three services measured by KPIs, specifically the Quality of Experience (QoE) for CV and Packet Loss (PL) for both URLLC and mIoT services. To maintain tractability, we have identified Packet Priority and Maximum Bit Rate (MBR) as the two key control parameters for each service. When the resources within the network slice are limited, increasing Priority/MBR for CV can enhance the QoE for CV service but may have an adverse effect on PL for URLLC and mIoT. Likewise, boosting Priority/MBR for URLLC can improve PL for URLLC but it may impact the PL for mIoT and QoE for CV services. In essence, the challenge lies in optimizing the fulfillment of these conflicting objectives, \textit{in accordance with given priorities}, while adhering to the resource constraints of the network slice. To address this challenge authors relied on MARL technique\cite{paper:globecom} where one MARL system (of atleast 3 RL agents) optimized packet priority for all three services while the other system was assigned to adjust MBR. While this method allowed autonomous execution for optimizing conflicting intents, it relied on a rule-based supervisor agent for orchestration of constituent MARL systems. A recent work \cite{paper:dey2023goals} extended this concept to introduce an AI-based supervisor agent using AHT approaches thereby alleviating the need for framing human-coded rules. 
However, the technique still needs a pre-defined utility function and priorities of each intent, \textit{provided apriori during training}, and cannot adapt to changes in priority or utility function(fig \ref{fig:intropic}). Thus it lacks the flexibility and adaptability for implementation in live networks.


\section{Proposed Methodology}
\label{sec:prop}

As part of this work, we propose a method(fig. \ref{fig:ICC_BD}) to implement a supervisor agent which is generic to the form of utility and can be flexibly redefined during execution. The supervisor agent uses a RL policy ($\pi_g$) which assigns sub-goals to two lower-level systems \textit{(i) Priority MARL} and \textit{(ii) MBR MARL system}. Each MARL system contains three RL agents, one agent for each instance of a service. Later to demonstrate scalability of our method, we extend the action space to include two new systems \textit{(iii)Move UE Context MARL system} and \textit{(iv)Auto Scale} agent for handling new expectations on the intents as part of Experiment 5 in section \ref{sec:results}.

The supervisor module, including the RL policy, can be trained using four types of inputs (i) Capabilities of agents as vector $\Gamma_{t}^i$ to Encoder (ii) Current performance of the each agent (tuple $s_t^i$,$a_t^i$,$g_t^i$),  where $s_t^i$ the current state of agent $i$, $g_t^i$ is the goal assigned to agent $i$ and $a_t^i$ the action by agent $i$ at time instant $t$.  (iii) Global Goals or Intents and (iv) Utility function $Z(\textbf{x})$, defining the intent fulfillment (embedded as Reward function in $\pi_g$). 
In \cite{paper:dey2023goals}, authors implemented the supervisor module which can ingest the first three inputs during the training and execution phase while the utility function was fixed and pre-defined as a part of RL policy $\pi_g$. In this work, we extend the framework by adding a DUN along with specific feature engineering \eqref{eqn:feature}. The novelty of the technique is the simplicity of the feature engineering and DUN, both of which requires minimal computational overhead for training of the supervisor agent. As a result, \textit{the proposed method enables the goal policy ($\pi_g)$ to generalize and adapt to various form of functions or priority values of intents even in execution phase}. 

During training phase, we train the policy ($\pi_g$) to generate sub-goals for each agent in MARL system using an Actor-Critic method.
The gradient of the Actor-critic is:
\begin{align}
 \; \nabla_{\theta}  J(\theta) \approx \frac{1}{n} \sum_{i} \nabla_\theta \: log \, \pi_\theta (\textbf{u}/S_G^i) \; \hat{A}^\pi (\textbf{u}/S_G^i) 
\end{align}where 
\begin{align}
    \hat{A}^\pi (\textbf{u}/S_G^i) = r(S_G,\textbf{u}) + \; \gamma\, \hat{V}^\pi_\phi(S_{G}^{'}) -  V^\pi_\phi(S_G)
\end{align}  
Here $r(S_G,\textbf{u})$ is the global reward that IMF accrues due to joint actions of lower level MARL systems and $S_G$ refers to the global goal-conditioned state\cite{nasiriany2019planning} and $\textbf{u}$ the actions (sub-goals for agents) given by supervisor goal policy ($\pi_g$) respectively. $\hat{A}^\pi$ refers to the advantage function of the critic network. The global reward at each time-step is calculated from equation \eqref{eq:eq_obj}. Based on the accrued reward signal the goal policy ($\pi_g$) is obtained during the training phase.

Given the utility function (\ref{eq:eq_obj}) is fixed during training, we propose a method whereby \textit{features of the user-defined utility function is passed through a network (\textit{DUN})} thereby giving the policy network ($\pi_g$) an ability to infer the function characteristics. Since such features can be modified during execution time, this gives the method a mechanism to tune its behavior ($\pi_g$) as feature vector changes. 
For this we send the current KPI values $x_i$ values for expectation $i$ to a "Feature Engineering" module to output the features $y_i$ as 

\begin{equation}
y_i= P_i \cdot \frac{f(|KPI_{current} - KPI_{target}|)}{f(Range \, of\, the\, KPI)}
\label{eqn:feature}
\end{equation}

where $P_i$ denotes the priority of the intent. The input feature to \textit{DUN} is normalized by range of the function $f(.)$ to provide an indication to \textit{DUN} about the relative importance of the current priority given the KPIs of QoE and Packet Loss are measured in different scales. 
The \textit{DUN} output context $n_{t+1}$, is fused with the state-performance embedding of the agent ($m_t^i$) and with Global Goals (intents) through Fusion Layer. The Fusion Layer outputs a context vector $c_{t+1}$. This global context is used to train the goal policy $\pi_g$ which outputs the sub-goals of constituent agents for next time-step ($t+1$). The sub-goals help in orchestration of the MARL systems autonomously. The Fusion layer and the goal policy is trained together with \textit{DUN} and the gradients are back propagated synchronously. 

During training, we train $\pi_g$ using the linear form of $f(.)$ as shown in \eqref{eq:lin} for all intents. We test it under logarithmic and quadratic forms as shown in equations \eqref{eq:log} and \eqref{eq:square} respectively with changed priorities. It may be noted that during execution, no reward signal is available but the pre-trained policy ($\pi_g$) is able to generalize to any new set of priorities or to unseen function forms leveraging the feature context from \textit{DUN}. Hence the retraining of policy ($\pi_g)$  is \textbf{not} required.

In experiments, the Encoder is a 2-layer Fully Connected Network followed by Merger layer which is also a 2-layer network. The DUN is chosen to be 2-layer connected network and Fusion layer used in this work is a 3-layer network. As for the Actor-Critic, the actor is a 2-layer Gated Recurrent network(GRU)\cite{paper:GRU} and critic is a 2-layer network. 

As a baseline, we have compared the proposed approach with the method in \cite{paper:dey2023goals} and have used IAE as a comparison metric \eqref{eq:IAE}. In both cases the model is trained with Linear utility and evaluated on Logarithmic and Squared utility functions. 

\section{Results and Discussions}
\label{sec:results}

To test the proposed approach we have used the network emulator in \cite{paper:globecom} which has capability to run three services (i) CV, (ii) URLLC and (iii) mIoT. These services are hosted on application servers which can send/receive data to/from UE's through gNodeB's and UPF's. The emulator also has the flexibility to change the airlink bandwidth (between gNodeB to/from UE's) to create scenarios like enough or scare resource situations. \textit{We used two values for airlink bandwidth in our work (i) 20 MBPS which reflect the situation where there are just enough resources to achieve the intents and (ii) 10 MBPS, reflects the situation that there is \textbf{not} enough of resources to achieve all the expectations given by the customer intents.}


We seek to optimize three intents in this work. The first intent is on CV service for maintaining an expectation on QoE $\geq$ 3 for UE's accessing CV, second intent is on URLLC for maintaining expectation on PL $\leq$ 2\%  and then on the mIoT service for $\text{PL}\leq$ 4\%. In later part we also add three new expectations: 1) \textit{Latency} for URLLC intent 2) \textit{Latency} for mIoT intent and 3) Minimization of \textit{Power Consumption} for mIoT intent which is in line with global sustainability goals. 

We have considered three forms of the utility function here. The function form for intent(or expectation) $i$ is defined as  
\begin{align}
    \text{Form 1}: f_i(x) = -P_i(|KPI_i - Target_i|) \label{eq:lin}\\
    \text{Form 2}: f_i(x) = -P_i(\ln(|KPI_i - Target_i)) \label{eq:log}\\
    \text{Form 3}: f_i(x) = -P_i(KPI_i - Target_i)^2 \label{eq:square}
\end{align}

where $P_i$ corresponds to assigned priority for an intent(or expectation/s in the intent). Note that utility function and Priority \textit{can be} defined at granularity of an "expectation" but when an intent consists of one expectation, both definitions are synonymous. Hence they have been used interchangeably in some instances. 
To explain the importance of the utility functions, keeping $P_i=1$ for all intents, for example, when net deviation from target($KPI_i - Target_i$) changes from $7$ to $5$, it results in net change of $2$ for form $1$, $0.33$ for form $2$ and $24$ for form $3$. This translates to a stringent expectation on the IMF for the intent driven by quadratic function. Consequently the KPI fulfillment and convergence of goals for the form 3 needs to be prioritized by the IMF.
Given the need for intent fulfillment while maximizing resource optimization, operators in future, will need to leverage different forms for utility functions in order to \textit{influence differential IMF behaviors based on monetization potential of the intent}. 


\begin{figure*}
    \centering
    \subfloat[Testing of baseline on Scenario 1]{\includegraphics[scale=0.3]{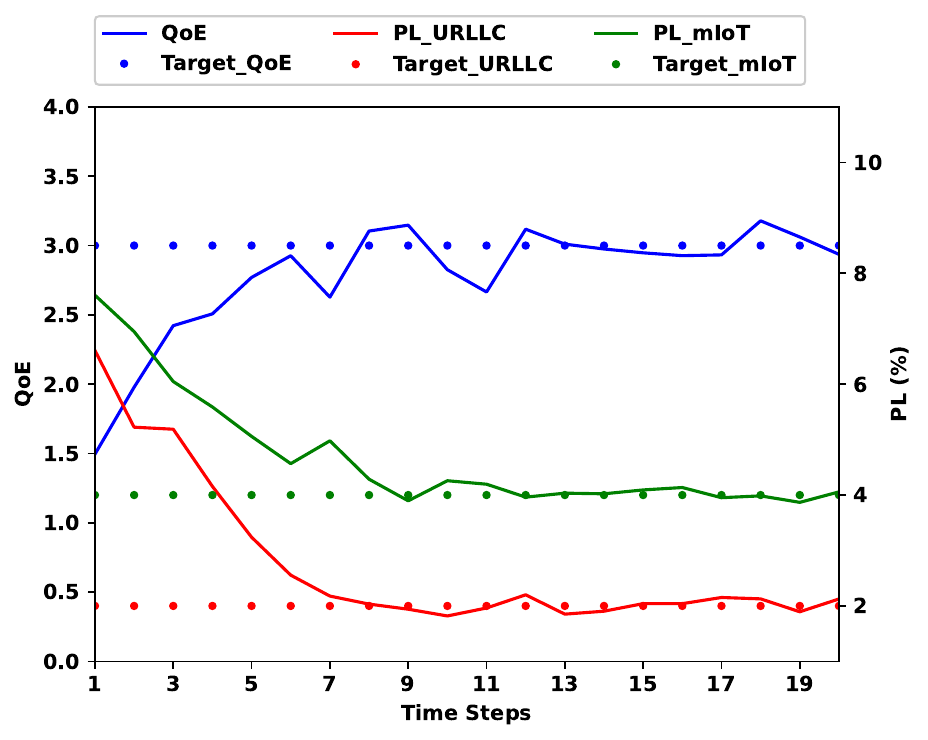}\label{fig:First_Baseline_Convergence_Scratch}}
    \subfloat[Testing of baseline on Scenario 2]{\includegraphics[scale=0.3]{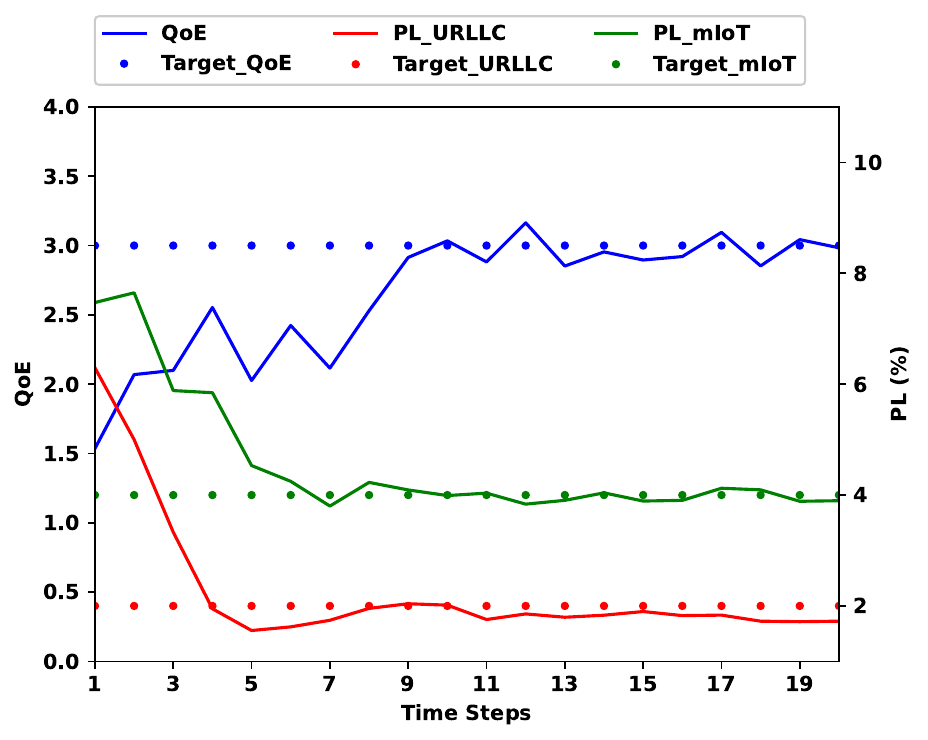}\label{fig:Second_Baseline_Convergence_Scratch}}
    \subfloat[Testing of baseline on scenario 2 using agents trained on Scenario 1]{\includegraphics[scale=0.3]{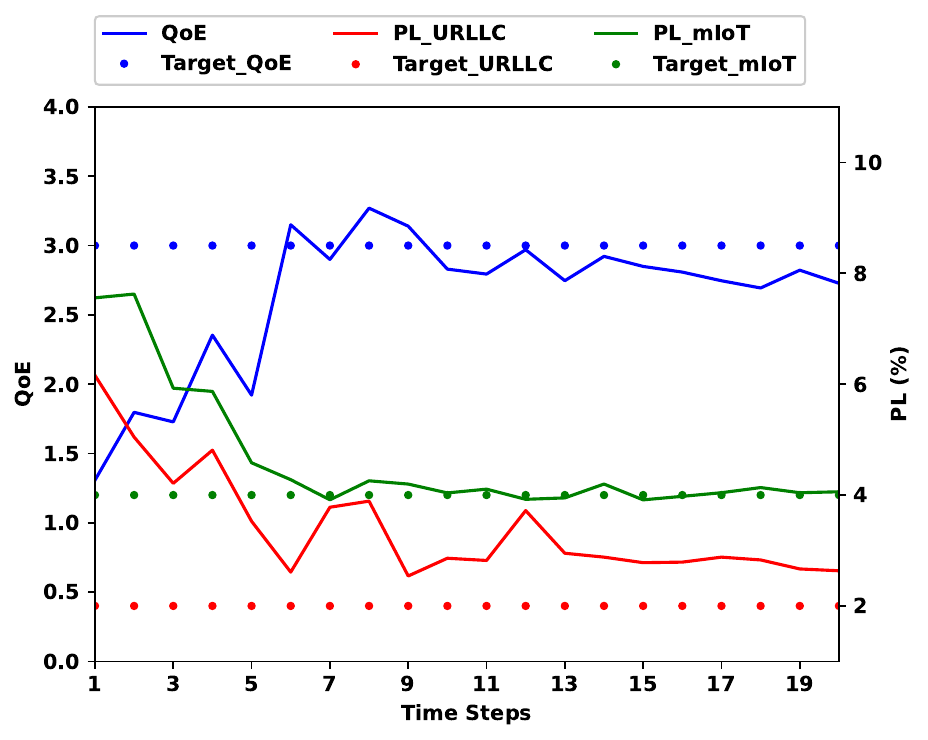}\label{fig:Linear_Baseline_Convergence}}\\
    \subfloat[Testing of proposed method on Scenario 1]{\includegraphics[scale=0.3]{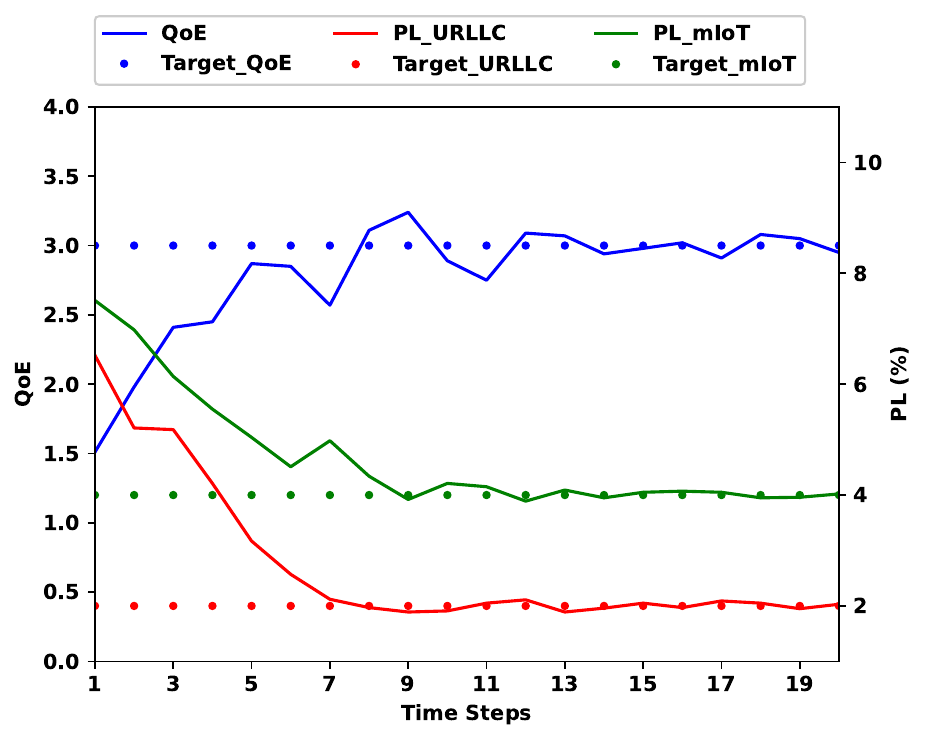}\label{fig:Linear_Linear_Convergence}}
    \subfloat[Testing of proposed method on Scenario 2]{\includegraphics[scale=0.3]{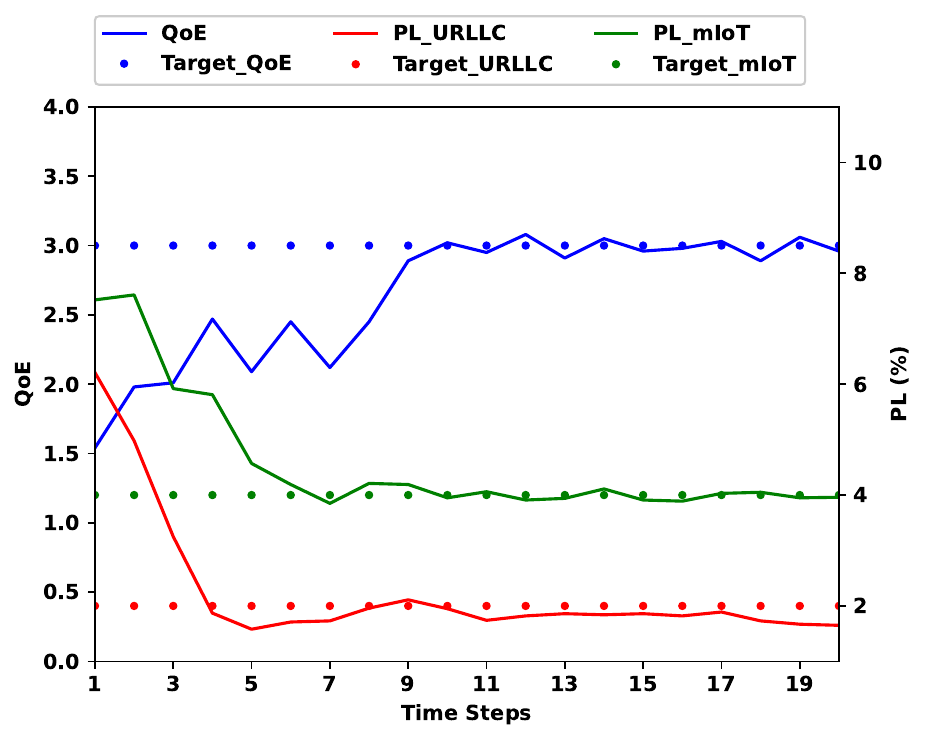}\label{fig:Linear_Utility_Convergence}}
    \subfloat[Testing proposed method on scenario2 using agents trained on Scenario 1]{\includegraphics[scale=0.3]{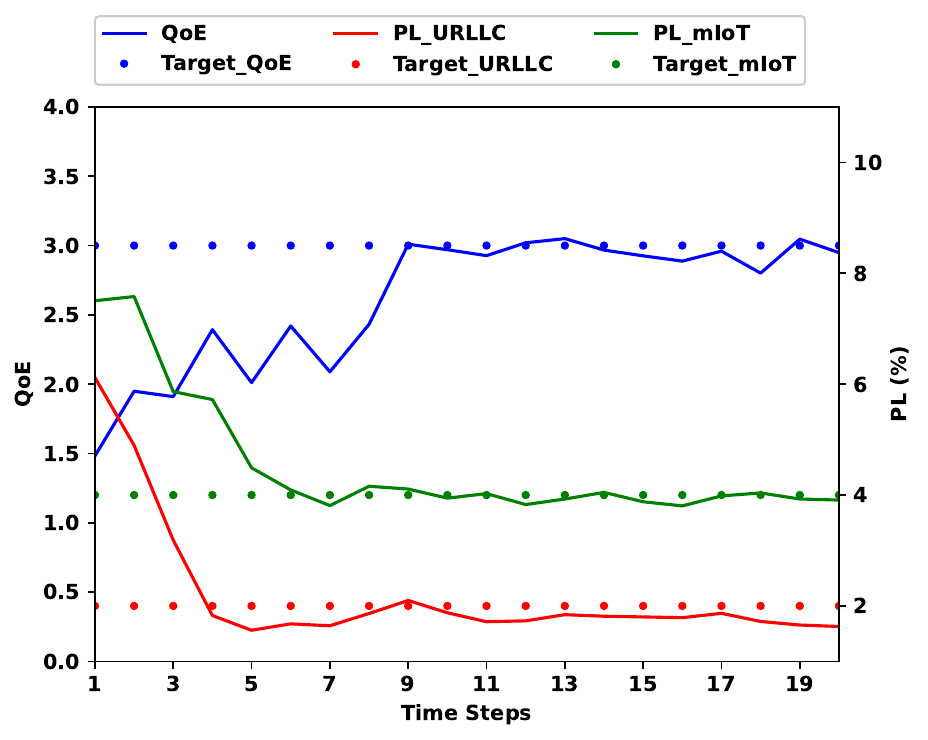}\label{fig:Linear_Proposed_Convergence}}
    \caption{Execution results of supervisor agent trained using baseline method (a,b) and proposed method (d,e) on scenarios 1 and 2. Both methods perform equally well in execution phase as long as priority and utility is not changed. Plot (c) corresponds to testing the baseline model trained in scenario 1 on scenario 2. Plot (f) corresponds to proposed method where baseline model trained in scenario 1 is tested on scenario 2.  From the plots, it is evident that existing baseline approach fails to generalize well whereas the proposed approach far outperforms the existing with change in priority and utility function. }
\end{figure*}


To investigate the impact of utility function on convergence speed, we conducted experiments using the scenarios in Figure 1, simulating ample resources, i.e 20 MBPS airlink bandwidth, to fulfill all intents.
We first train the priority and MBR agents, independently, as discussed in \cite{paper:globecom}. Then we trained the supervisor agent on \textit{Scenario1} and \textit{Scenario2} using baseline and proposed methods. A sample execution episode using the trained models is shown in fig. \ref{fig:First_Baseline_Convergence_Scratch}, fig. \ref{fig:Second_Baseline_Convergence_Scratch} for baseline and fig. \ref{fig:Linear_Linear_Convergence}, fig. \ref{fig:Linear_Utility_Convergence} for proposed methods on Scenario1 and Scenario2 respectively.
In scenario1, all the KPI's reach their respective targets around same time in both methods. However, in the second scenario, URLLC service KPI $PL_{urllc}$, subjected to quadratic utility(Form 3), converges earlier (fig.\ref{fig:Second_Baseline_Convergence_Scratch}, fig.\ref{fig:Linear_Utility_Convergence}) than the CV service KPI $QoE$ and mIoT service KPI $PL_{mIoT}$. Also, since the priority accorded is higher for mIoT, it converges faster than the CV service for both methods. 
Hence, it can be concluded that the a polynomial utility function can induce high priority and faster convergence. Same is true for higher priority value($P_i$) but the effect is less pronounced. Also comparison of fig. \ref{fig:Linear_Baseline_Convergence} with fig. \ref{fig:Linear_Proposed_Convergence} demonstrates the efficacy of generalization capabilities which in subsequent sections, is elaborated with additional experiments.

In order to demonstrate generalization of proposed approach for different combinations of utility function forms and $P_i$s, a large number of scenarios is experimented upon. However given the limitations of space, it is not feseable to show test results for all the cases. Hence, we have chosen to use a metric "IAE" (Integral Absolute Error) to quantify the generalization performance of the proposed framework, a metric which is widely used in closed-loop literature. The IAE is measured as
\begin{align}
    IAE = \frac{1}{N} \displaystyle \sum_{k=1}^N \frac{|KPI[k]-T|}{T}
    \label{eq:IAE}
\end{align}

where $KPI[k]$ is the value of the KPI at time instant $k$, $T$ is the value of target and $N$ is the length of the execution time period (20 time-steps in our case). 

The IAE value indicates the average value of deviation for $N$ time steps. If the value of IAE is small, it suggests that average deviation of KPI is smaller which indicates faster convergence and vice-versa. Hence, IAE value is considered as a good measure of speed of convergence and each point in the IAE graph corresponds to a convergence graph. 

 We have followed the procedure in \cite{paper:globecom} to train the Priority MARL system and MBR MARL system, each with three agents(corresponding to each service) for experiments 1-4 and have incorporated two new systems(fig. \ref{fig:ICC_BD}) for experiment 5. The following experiments are performed with an airlink bandwidth of 10 MBPS. This simulates a scenario which does not have sufficient resources to fulfill all intents and hence prioritization is needed. Then with pre-trained MARL systems in lower level, we have employed the proposed approach in Section \ref{sec:prop} to train the supervisor module \textit{using the Form $1$ of the utility function with a $P_i = 1 \;\;\forall\;\; \text{intents}$.} For comparison, we also trained the baseline method as in \cite{paper:dey2023goals} with form $1$.

At test-time we change the form of the Utility function to Form 2 \eqref{eq:log} and Form 3 \eqref{eq:square} respectively and evaluate our proposed method and compare to the current baseline.

\textbf{Experiment 1-}\textit{ Changing function form to log and intent $P_i$ at run time} : The supervisor is trained for linear intents and $P_i=1$ for all intents. During execution, the the utility function is changed to Form 2 \eqref{eq:log} for all intents, which was unseen at training time. 
Then the priority $P_i$ of the intent is monotonically increased from 1 in steps of 2 (incl. 10), keeping $P_i$ of other intents constant at 1. If the proposed method performs well to change in function form, we would expect the chosen service intent to steadily improve its IAE values as priority increases. This is exactly observed in Figure \ref{fig:Log_CV}, where the IAE value for CV continues to improve(decrease) while the IAE values for URLLC and mIoT degrade (increase) as $P_{CV}$ is increased in steps. The proposed algorithm is able to interpret the change in priorities, even after change in function form, and self-adjusts by allocating more network resources to CV taking the same from URLLC and mIoT.

For efficacy comparison, we refer to the baseline method (dotted lines in fig. \ref{fig:Log_CV}). The change of function form has rendered the baseline algorithm ineffective, whereby the prioritization of CV and deprioritization of URLLC and mIoT is not effected and the IAE value keeps oscillating. Similar effect can be observed in (Fig. \ref{fig:Log_URLLC} and \ref{fig:Log_MIOT}) for monotonic increase in URLLC and mIoT priorities respectively. 
This confirms that our proposed method is able to generalize to (i) Logarithmic function form and (ii) change in priorities, even though it was trained on linear form and on a fixed set of priority values.

\textbf{Experiment 2-} \textit{Changing function form to quadratic and intent $P_i$s at run time}:  Since the used function form is square \eqref{eq:square}, it amplifies the value and hence we can see a pronounced prioritization and consequently better performance for prioritized service (fig. \ref{quadratic}) when compared with logarithmic case, as $P_i$ value increases. If we compare the CV service, with $P_{CV}=8$ it is evident that the IAE values have improved more than $100\%$ in fig:\ref{fig:Square_CV} as compared to fig. \ref{fig:Log_CV}. Given this is an expected motivation of using the quadratic function over a logarithmic, and the same is observed, we infer the method generalises well to exhibit characteristics of chosen function.

The IAE values obtained for different values of $P_i$ of CV, URLLC and mIoT services, when $P_i$ values of other two services are fixed at $1$, is shown in fig. \ref{fig:Square_CV}, \ref{fig:Square_URLLC} and \ref{fig:Square_MIOT} respectively. The plots suggest improved generalization of the proposed approach. Comparably, the baseline fails to prioritize the intended service through necessary resource attribution. 

\textbf{Experiment 3- }\textit{Combination of forms of function}: In previous two cases, we assumed every intent will have the same form of utility function. However, in real life each intent may have a form of utility which is different from the other. To demonstrate this we considered CV intent to have a linear form of utility, URLLC intent- Square and mIoT- Log form. 

Leveraging the model trained in form 1 earlier, the IAE values obtained for this case for different values of $P_i$ for a service, where the $P_i$ value of other two services is fixed at $1$, is shown in the fig. \ref{fig:Comb_CV}, \ref{fig:Comb_URLLC} and \ref{fig:Comb_MIOT} respectively. Here also the expected prioritization in seen in IAE for CV, URLLC and mIoT in the respective figures. For example, when $P_{URLLC}$ is increased in fig:\ref{fig:Comb_URLLC}, URLLC is seen to achieve high degree of fulfillment as compared to when $P_{mIoT}$ is increased in fig:\ref{fig:Comb_MIOT}. This happens because of extreme prioritization of quadratic function form assigned to URLLC intent vs the log form assigned to mIoT intent. This might be practically important in real scenarios where a real-time URLLC operation might need to be prioritized than an IoT application.

Hence with above three experiments, it may be concluded that our proposed method can \textit{generalize to various combination of utility function forms and adapt to changing priority (or both) without additional training} and thereby far outperforms existing SOTA for generalization in IMFs.
It might be good to mention that fixing the priority of other two intents is not necessary as we increase the priority of the chosen intent. It is done to demonstrate the resource attribution more distinctly as changing all three variables can make analysis more complex.

\begin{figure}
    \centering
    \includegraphics[scale=0.3]{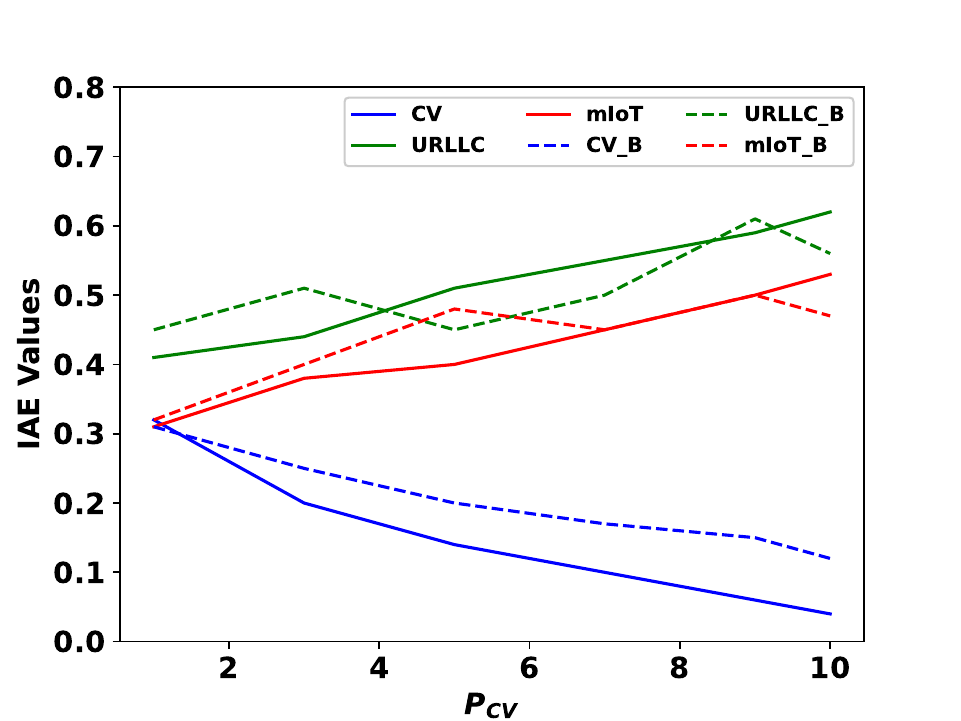}
    \caption{IAE values of all the KPIs vs $P_i$ for CV with $P$ at $1$ for URLLC and mIoT. No change in function form. }
    \label{fig:P_change}
\end{figure}
\begin{figure*}
\centering
\subfloat[IAE values as $P_{CV}$ changes]{\includegraphics[scale=0.3]{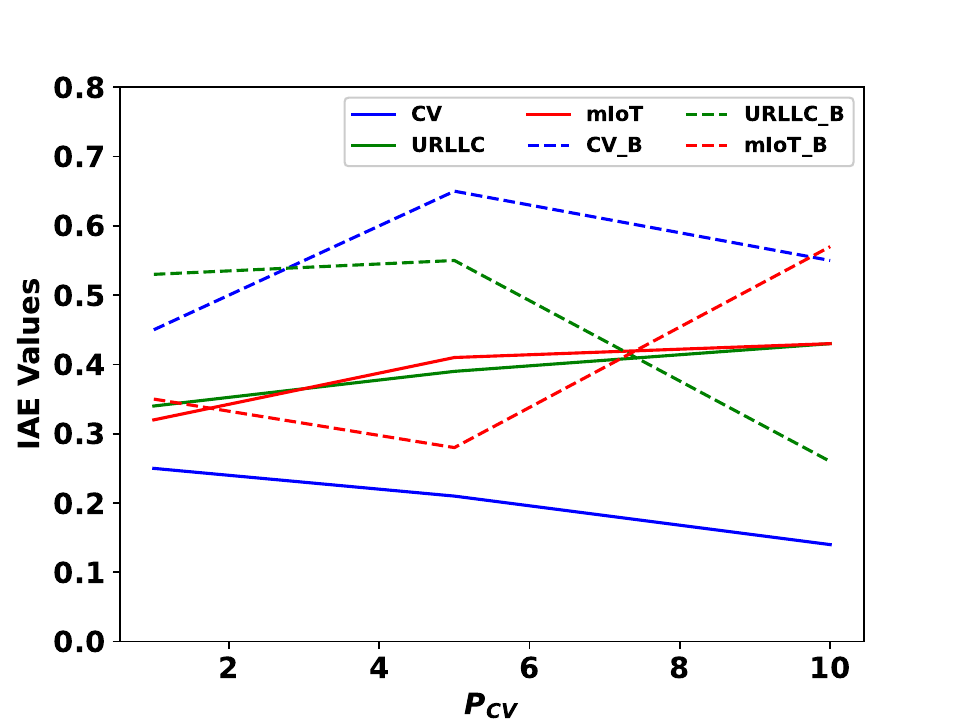}\label{fig:Log_CV}}\hfil
\subfloat[IAE values as $P_{URLLC}$ changes]{\includegraphics[scale=0.3]{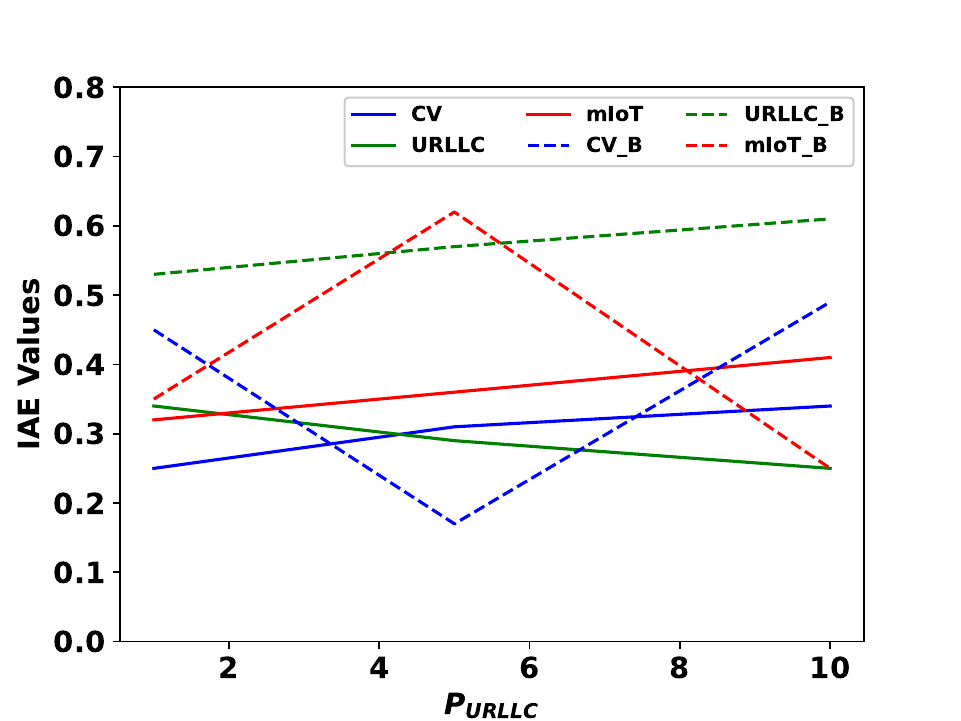}\label{fig:Log_URLLC}}\hfil
\subfloat[IAE values as $P_{mIoT}$ changes]{\includegraphics[scale=0.3]{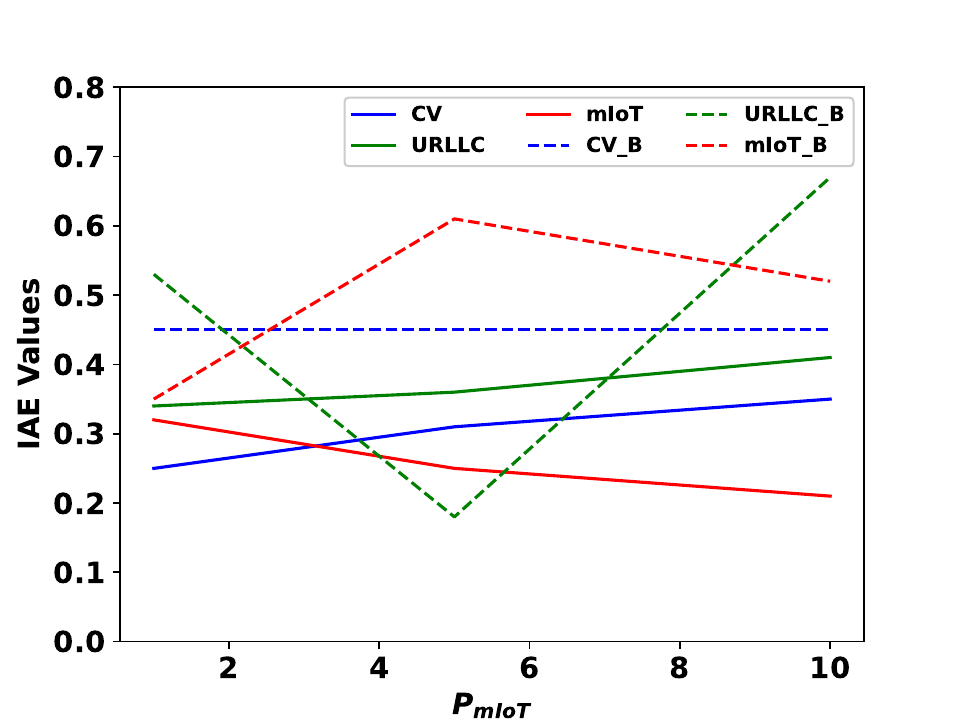}\label{fig:Log_MIOT}}\hfil
\caption{Evaluation on Form2 \eqref{eq:log} of utility function: Variation in IAE values of all KPIs vs $P_i$ values for prioritized intent keeping $P_i$ for other intents fixed at $1$. Solid lines correspond to proposed approach whereas dashed lines are the baseline. }
\end{figure*}
\begin{figure*}
\centering
\subfloat[IAE values as $P_{CV}$ changes]{\includegraphics[scale=0.33]{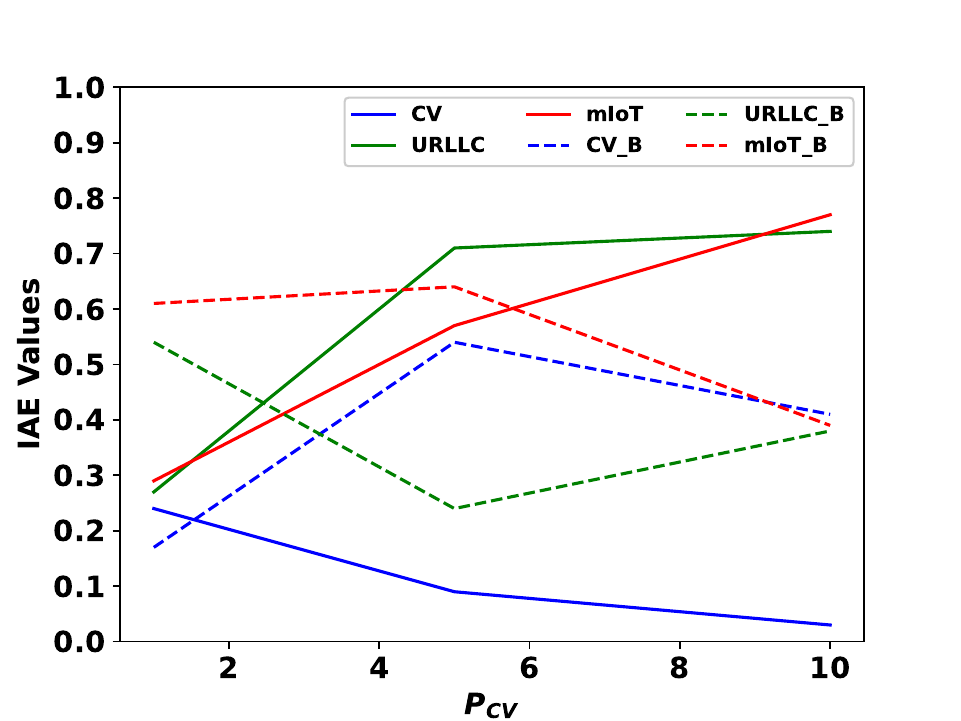}\label{fig:Square_CV}}\hfil
\subfloat[IAE values as $P_{URLLC}$ changes]{\includegraphics[scale=0.3]{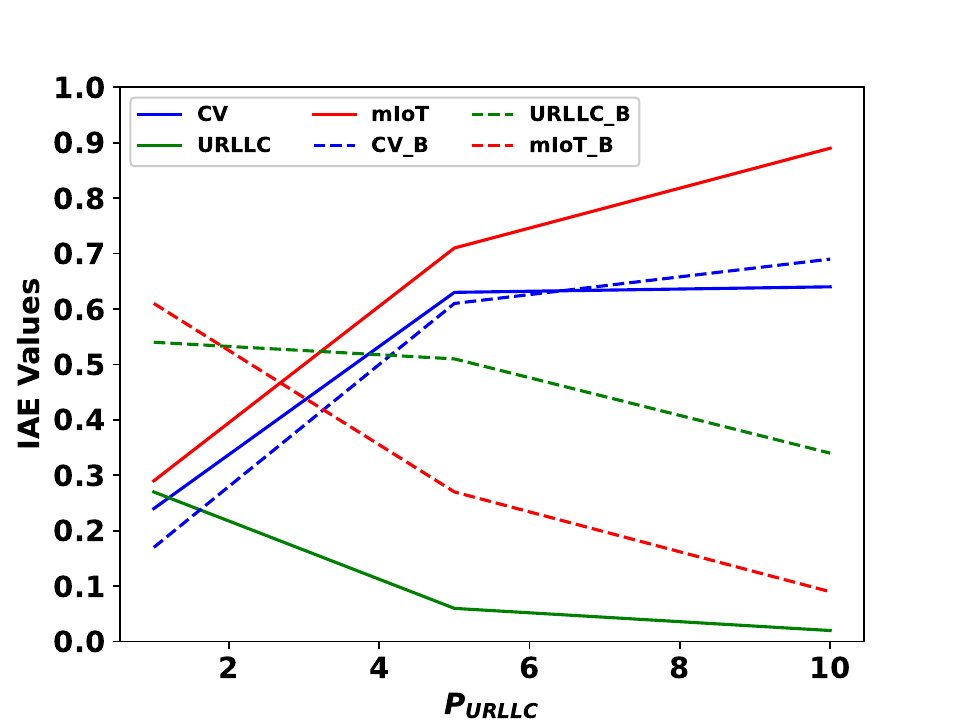}\label{fig:Square_URLLC}}\hfil
\subfloat[IAE values as $P_{mIoT}$ changes]{\includegraphics[scale=0.3]{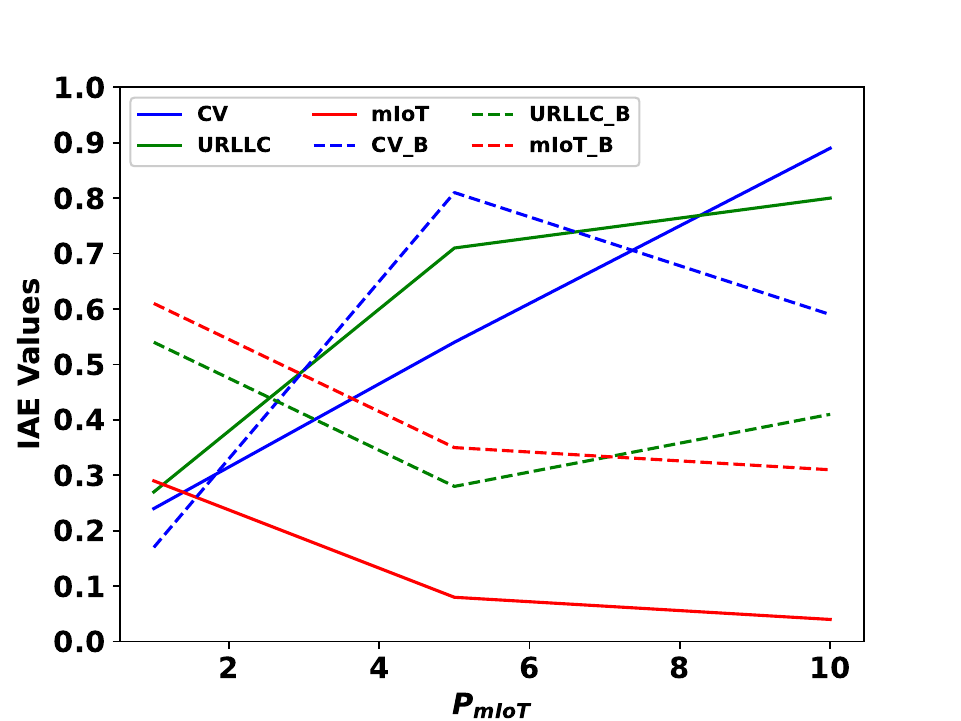}\label{fig:Square_MIOT}}\hfil
\caption{Evaluation on Form3 \eqref{eq:square} of utility function: Variation in IAE values of all KPIs vs $P_i$ values for prioritized intent keeping $P_i$ for other intents fixed at $1$. Solid lines correspond to proposed approach whereas dashed lines are the baseline.}
\label{quadratic}
\end{figure*}
\begin{figure*}
\centering
\subfloat[IAE values as $P_{CV}$ changes]{\includegraphics[scale=0.3]{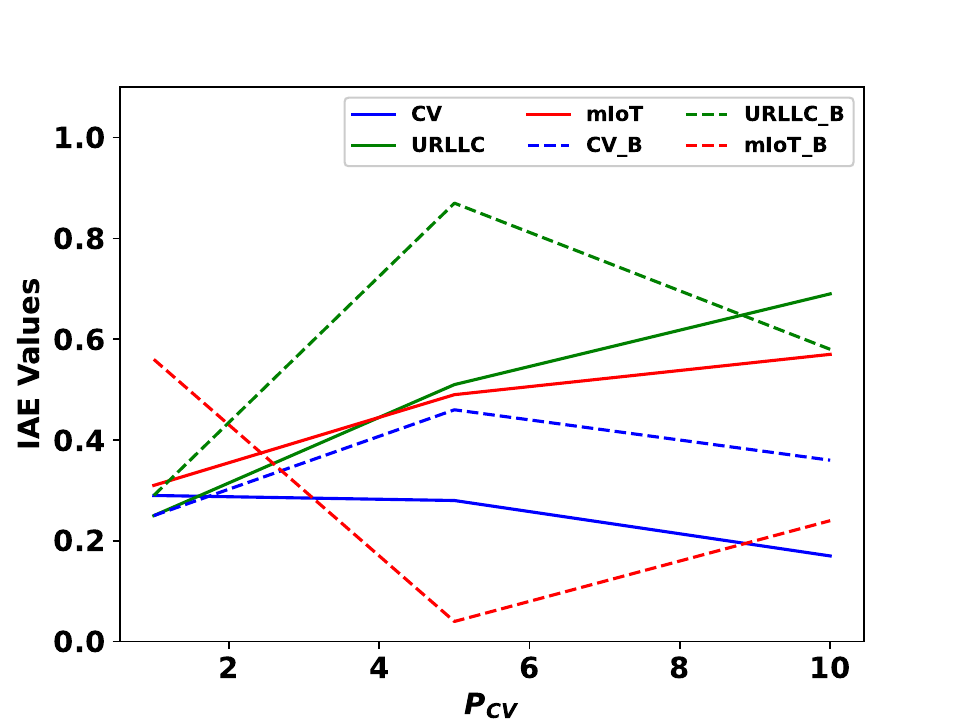}\label{fig:Comb_CV}}\hfil
\subfloat[IAE values as $P_{URLLC}$ changes]{\includegraphics[scale=0.3]{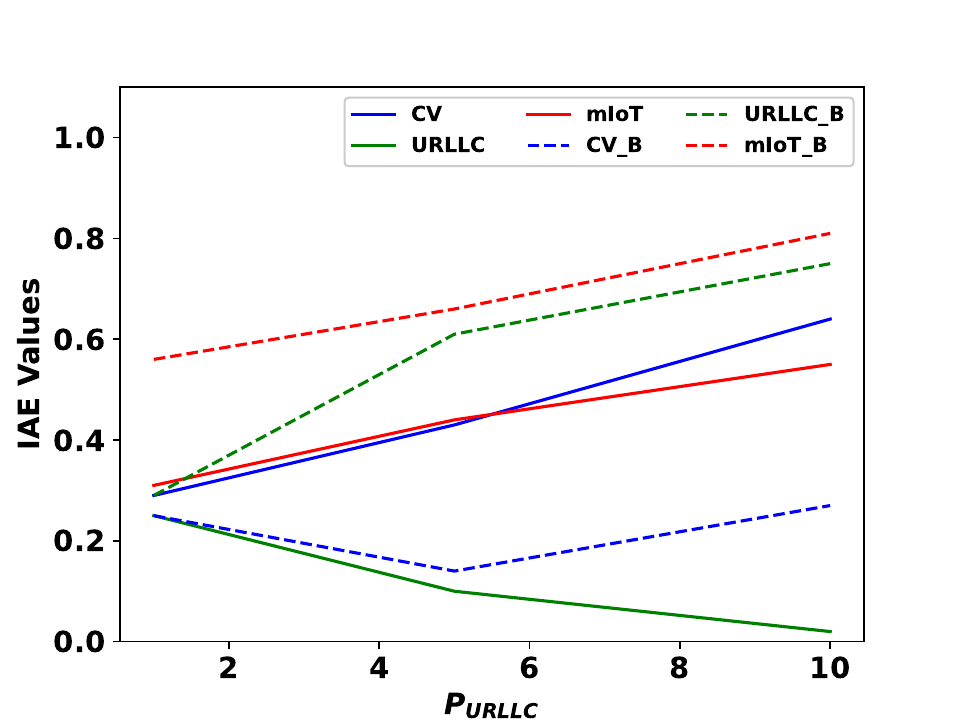}\label{fig:Comb_URLLC}}\hfil
\subfloat[IAE values as $P_{mIoT}$ changes]{\includegraphics[scale=0.3]{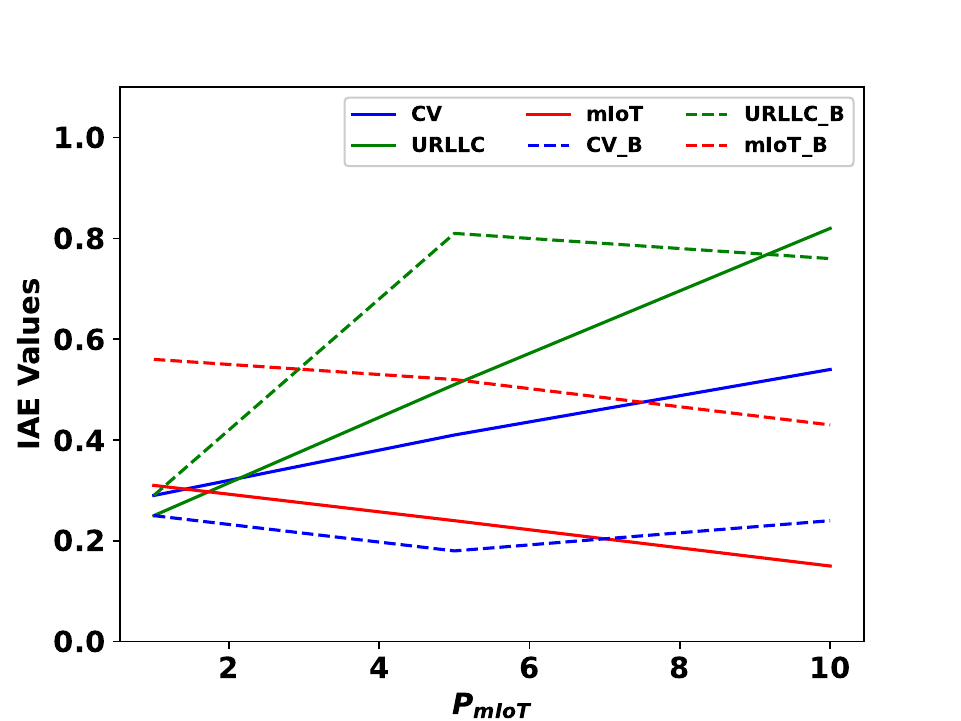}\label{fig:Comb_MIOT}}\hfil
\caption{Evaluation on Combining forms of utility function:  IAE values of all KPIs vs $P_i$ values for prioritized intent keeping $P_i$ for other intents fixed at $1$. CV tested on linear \eqref{eq:lin}, URLLC on Quadratic \eqref{eq:square} and mIoT on Log \eqref{eq:log} form }
\end{figure*}

\textbf{Experiment 4 -}\textit{Changing priorities at run-time while keeping function form as Linear}: Finally, we conduct an ablation experiment. Keeping the function form to linear for all intents, only the priorities of a chosen service is changed from $1$ to $10$, keeping other two service priorities constant at $1$. For constraints of space, we only show the CV prioritization. As evident in fig. \ref{fig:P_change}, with increase in CV priority the IAE for CV decrease with consequent increase in IAE of URLLC and mIoT. Our proposed method also outperforms the baseline. However, interestingly, the baseline method performs better in this simpler case than compared to cases where function form changes along with priority. This confirms, the more complex the change in function, more significant would be the need to implement our proposed method. Similar results are observed with URLLC and mIoT prioritizations on linear form.

\begin{figure*}
\centering
\subfloat[Baseline]{\includegraphics[scale=0.28]{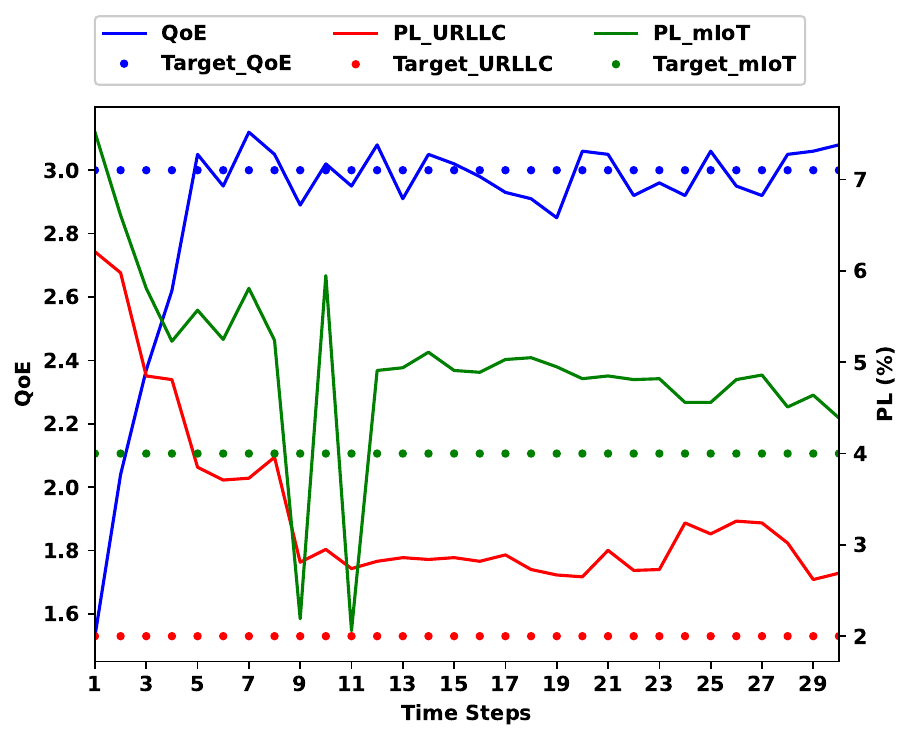}\label{fig:exp_5_base_QoE}}\hfil
\subfloat[Proposed Approach]{\includegraphics[scale=0.28]{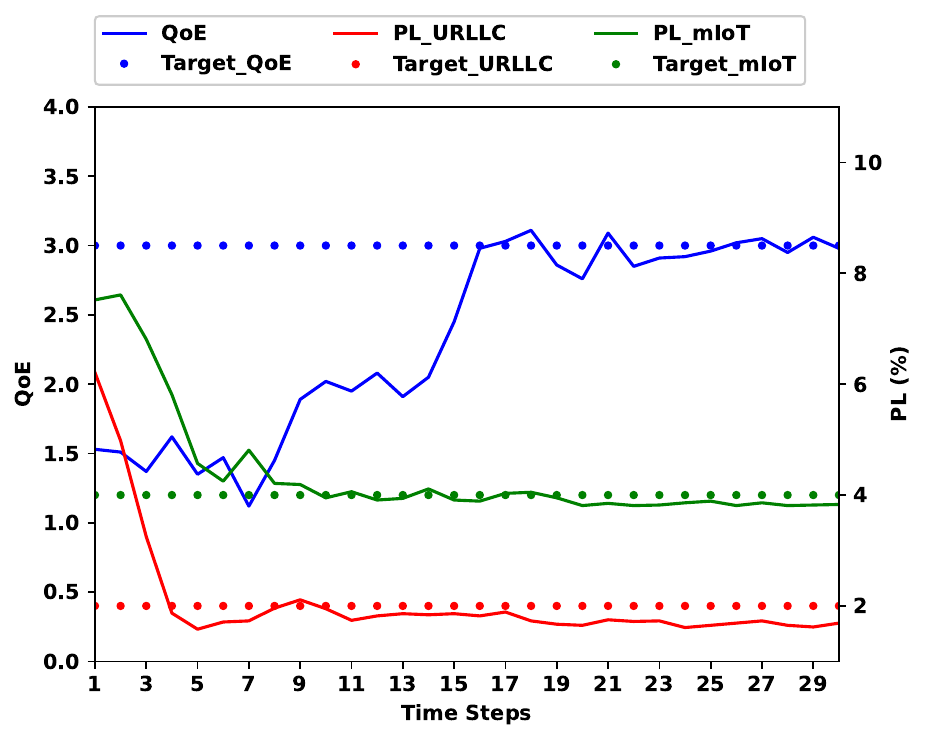}\label{fig:exp_5_prop_QoE}}\hfil
\subfloat[Baseline]{\includegraphics[scale=0.28]{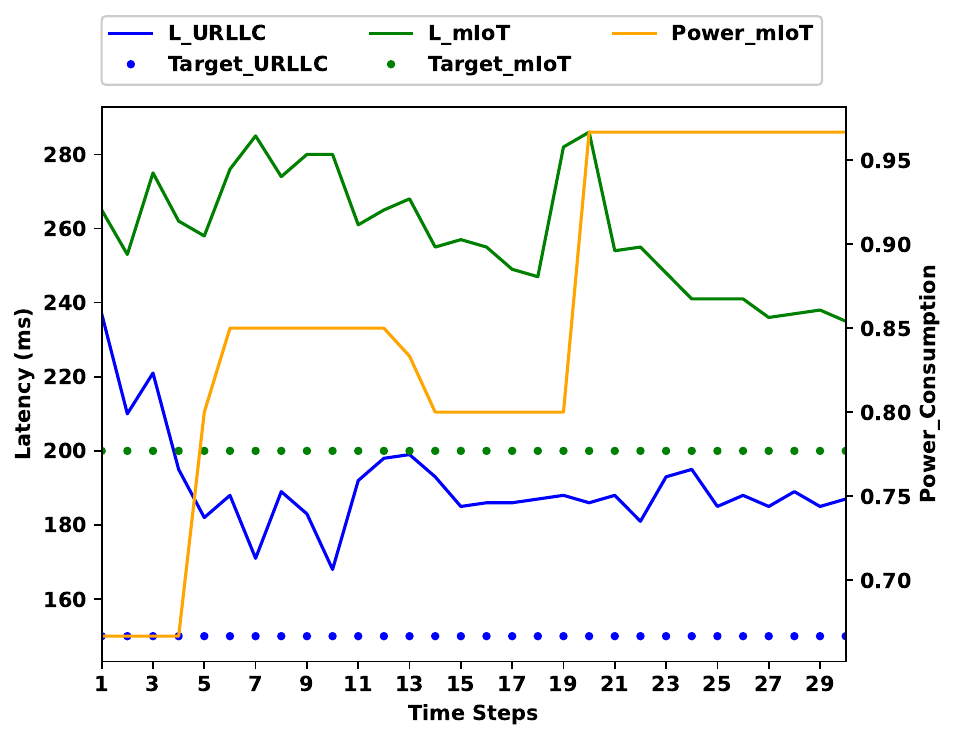}\label{fig:exp_5_base_Lat}}\hfil
\subfloat[Proposed Approach]{\includegraphics[scale=0.28]{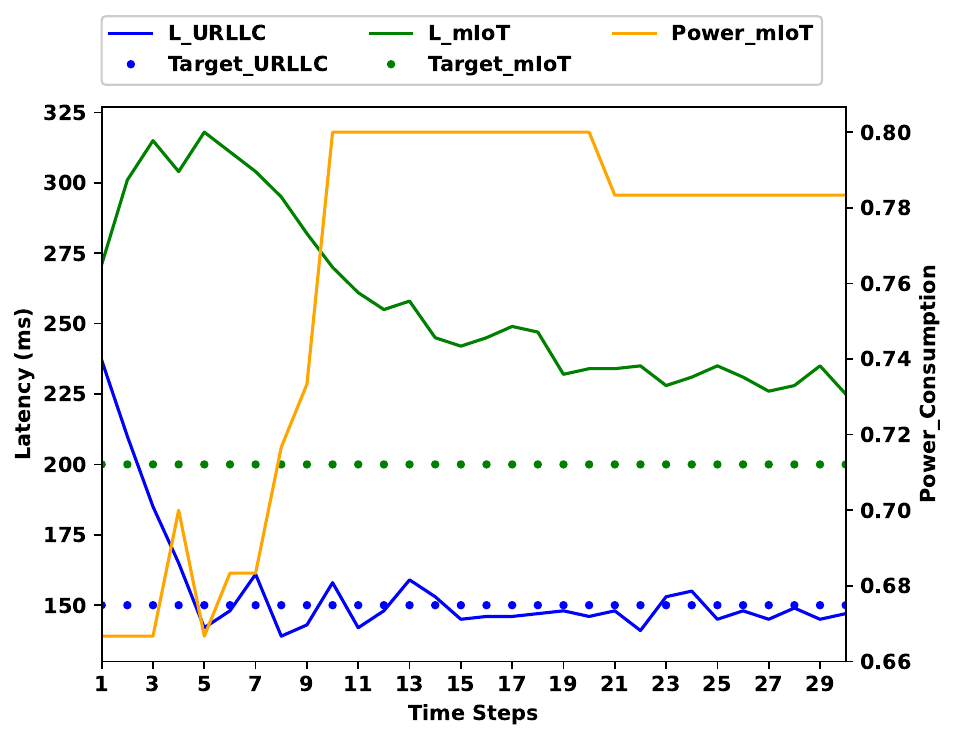}\label{fig:exp_5_prop_Lat}}\hfil
\caption{Evaluation results on scenario in Table II for proposed and baseline approaches. The proposed method achieved good generalization (b,d) whereas the baseline approach results in poor generalization (a,c). }
\label{fig:latency_fig}
\end{figure*}

\textbf{Experiment 5 -}\textit{Scalability of proposed method}: In literature\cite{paper:dey2023goals}, The IMF has been tested for scalability by increasing the instances i.e having more than 1 instance of an expectation on URLLC and mIoT services. This requires the system to work on same action spaces. However increasing action space and consequently having a new expectation on an intent has not yet been attempted. Notably, for intents on URLLC and mIoT, \textbf{Latency} is also an important expectation which needs to be optimized along with PL. An application may be deployed at multiple sites in the same time, and different UEs may access the same service from different sites. The Latency for URLLC and mIoT UEs can be reduced by moving the serving instances of applications for certain UEs(the UE context) from one node to another( for eg. from Central to Edge node). We refer to this action as \textbf{Move-UE-context}. Moving the UE context from central to edge node decreases (improves) latency and from edge to central increases (degrades) latency.
Similarly the serving node has a compute capacity in terms of number of CPUs or pods it can scale to. An \textbf{Auto-Scale Limit} is introduced which determines how many pods may be available to the serving application. The auto-scale limit is expressed in units of “small”, “medium”, “large”, "very large" and controls the number of pods the service is allowed to scale on a certain node. For e.g., a deployment with an autoscale limit of “large” implies a service can scale up to 8 pods. If an application is scaled to more pods, it can provide computations faster, thus this can be seen by the UEs as decreased end-to-end latency(NB: we define latency as measured by the client, which includes processing latency in the applications). Hence a combination of the actions \textbf{auto-scale limit} with \textbf{Move-UE-Context} can be leveraged by the agents to control the latency. 

\textbf{Power Consumption} is also an KPI for mIoT service given the large number of mIoT devices. Optimization of the same specially for mIoT services is deemed important for sustainability goals. We assume serving more UEs from an edge location may cost more energy as compared to a central site. This is due to central site having better provision for efficient compute resources and in edge it is more of an adhoc provisioning. For brevity we consider an mIoT UE served from central site consumes 50 units of power while that from two edge sites cost 60 and 70 units respectively in the emulator. The total power consumption is then normalized with possible maximum power consumption in a range of 0 to 1. Since our goal is to minimize the power consumption, a target of $0$  is given to encourage the system to reach minimum power configuration rather than reach a particular target. It is to be noted that while serving more UEs from edge improves latency but increases power consumption. Hence the agents are needed to do an appropriate trade-off. We exclude auto-scale limit as an possible action space for power consumption because of the limited effect of the action. We recognize there could are other relevant actions also but given the limitations of the emulator, our objective is to demonstrate the scalability of the framework for conflicting expectations. To summarize, CV is measured by QoE, URLLC by PL and Latency and finally mIoT intent by PL, Latency and Power Consumption.
The Table \ref{tab:table1} summarizes the Service KPIs and the corresponding action space:

\begin{table}[h!]
  \begin{center}
    \caption{Service, Expectation-KPI and Action Mapping}
    \label{tab:table1}
    \begin{tabular}{l|l|l|l}
      \textbf{Service} &\textbf{Expectation-KPI}& \textbf{Action 1} & \textbf{Action 2}\\
     \hline
      &    &\\
      CV   &QoE & Packet Priority    & MBR\\
      URLLC & Packet Loss & Packet Priority  & MBR \\
      URLLC & Latency &Move-UE Context & Auto-scale limit \\
      mIoT & Packet Loss& Packet Priority & MBR \\
      mIoT & Latency&Move-app & Auto-scale limit \\
      mIoT & Power Consumption& Move-app & - \\ 
     \hline
    \end{tabular}
  \end{center}
\end{table}

\begin{table}[h!]
  \begin{center}
    \caption{Evaluation scenario for 6 expectations}
    \label{tab:table2}
    \begin{tabular}{l|l|l|l}
      \textbf{Service-KPI} & \textbf{Target} & \textbf{Utility Form} & \textbf{$P_i$}\\
     \hline
      &    &\\
      CV - QoE   & $>=3 $ &Linear    &1\\
      URLLC - Packet Loss & $ <=2\% $ & Quadratic  & 1 \\
      URLLC - Latency & $ <=150 $ ms & Quadratic & 1 \\
      mIoT - Packet Loss  & $ <=4\% $ & Linear & 5 \\
      mIoT - Latency & $ <=200 $ ms & logarithmic & 3 \\
      mIoT - Power Consumption & 0 (min. possible)& linear & 1 \\ 
     \hline
    \end{tabular}
  \end{center}
\end{table}
This module includes 2 additional systems along with priority and MBR MARL systems, 1) A MARL system consisting of two agents, both modulating the move-UE-context action, one for Latency reduction another for optimizing power consumption 2) An single RL agent which can perform auto-scale actions in edge or central sites to change Latency in conjunction with move-UE-context action. Hence we scale the supervisor to coordinate 4 independent systems in order to achieve 6 expectations on same 3 services.

To begin with, all 6 expectations given in table \ref{tab:table1} are trained with linear utility function and $P_i=1$. 
Once trained, the trained framework is tested on a new scenario where the utility function and priorities are quite different as in Table \ref{tab:table2}

The results in fig.\ref{fig:latency_fig} demonstrates significantly improved generalization capability across all KPI expectations. At test time, the quadratic utility function even with smaller $P_i$ value (URLLC latency) achieves faster convergence when compared with Linear utility  with higher $P_i$ (mIoT PL) which in turn gets preference over logarithmic utility (mIoT Latency). The Power Consumption KPI performance also improves by $\approx 20\%$. 
This demonstrates that our proposed method can scale effectively even with increase in heterogeneity of action spaces.

\section{Conclusions}
\label{sec:conc}
The proposed method enables an AI-Based IMF with complex lower level agents, to generalize to utility function definitions at run-time. While the training can be done with a certain utility form, the method provides a way to change the function form and the priorities of an intent during execution. Results demonstrate the proposed method outperforms existing methods where such a generalization in IMF is not feasible without additional training. The proposed method can also scale when new types of action spaces and intents are introduced.
This is probably the first attempt at utility function generalization in a multi-service multi-intent IMF framework and would have significant impact in building \textbf{Adaptive Intent Based IMFs} for live implementations in future 6G networks.

\bibliographystyle{plain}
\bibliography{References}

\end{document}